\documentclass[11pt]{airesearch}

\usepackage{fullpage}
\usepackage{wrapfig}
\setlist[itemize]{leftmargin=*,topsep=1pt,itemsep=1pt}
\RenewDocumentCommand{\paragraph}{s o m}{%
  \par\medskip\noindent\textbf{#3}\quad\ignorespaces
}

\usepackage{algorithm}
\usepackage{algpseudocode}  

\usepackage{fancyhdr}
\newcommand{\firstpageheader}[1]{%
  \thispagestyle{fancy}%
  \fancyhf{}%
  \fancyhead[L]{\small \rmfamily #1}
  \fancyfoot[C]{\thepage}%
}

\newcommand{\projectpage}[1]{%
  \begin{center}
    \small
    \vspace{-1.25ex}
    \texttt{Project Page}: \url{#1}
  \end{center}
}

\newcommand{\metadata}[2][Metadata]{%
  \begin{center}
    \small
    \vspace{-1.25ex}
    \texttt{#1}: #2%
  \end{center}
}

\ifdefined\final
  \usepackage[disable]{todonotes}
\else
  \usepackage[textsize=tiny]{todonotes}
\fi

\title{\huge \bfseries \sffamily Training and Evaluating Language Models with Template-based Data Generation}
\author{\textbf{Yifan Zhang}\\University of California, Los Angeles\\ \texttt{yifzhang@ucla.edu}}
\date{}

\begin{document}
\maketitle

\begin{abstract}
The rapid advancement of large language models (LLMs) such as GPT-3, PaLM, and Llama has significantly transformed natural language processing, showcasing remarkable capabilities in understanding and generating language. However, a fundamental bottleneck persists: these models often struggle with tasks requiring complex, multi-step reasoning, particularly in mathematical problem-solving. This deficiency stems from the critical scarcity of large-scale, high-quality, domain-specific datasets necessary for cultivating sophisticated reasoning abilities. To overcome this challenge, we introduce \textbf{Template-based Data Generation} (\textbf{TDG}), a novel and scalable paradigm that harnesses frontier LLMs (GPT-4) to automatically generate parameterized meta-templates, which in turn synthesize a virtually infinite stream of high-quality problems and solutions. Using this paradigm, we create TemplateMath Part I: TemplateGSM, a foundational dataset of over 7 million synthetically generated grade school math problems. Each problem is accompanied by a programmatically verifiable solution, offering an unprecedented level of quality at scale. This resource not only resolves the data scarcity issue for supervised fine-tuning but also provides a robust mechanism for model alignment through Reinforcement Learning with Verifiable Rewards (RLVR). Our approach elevates data augmentation by leveraging GPT-4 to generate meta-templates, ensuring diverse and complex problem structures. By providing a scalable solution to the data and verification bottleneck, TDG and TemplateGSM pave the way for a new generation of LLMs with powerful, reliable reasoning skills.
\end{abstract}

\metadata[Dataset]{\url{https://huggingface.co/datasets/math-ai/TemplateGSM}}
\metadata[Project Page]{\url{https://github.com/iiis-ai/TemplateMath}}

\firstpageheader{Published in ICLR 2025 DATA-FM Workshop}

\section{Introduction}

Large language models (LLMs) have revolutionized natural language processing (NLP), exhibiting unprecedented capabilities in language understanding and generation. Models such as GPT-3 \citep{brown2020language}, PaLM \citep{chowdhery2022palm}, and Llama \citep{touvron2023llama} have achieved remarkable success across various NLP tasks, including machine translation, summarization, and question answering.

Despite these advancements, a critical frontier remains: enabling LLMs to perform complex, multi-step reasoning with precision and reliability. This capability is a cornerstone of advanced intelligence, yet models often falter in domains requiring it, most notably mathematics \citep{hendrycks2021measuring, cobbe2021training, wei2022chain}. Mathematical reasoning presents a critical challenge, requiring rigorous logic and exact solutions. The primary obstacle is not architectural but rather a data-centric one: the profound scarcity of large-scale, high-quality, and diverse mathematical datasets with verifiable solutions \citep{paster2023openwebmath, azerbayev2023proofnet}. This data bottleneck severely impedes the training, evaluation, and alignment of LLMs, hindering progress toward more capable and reliable foundation models.

To fundamentally address this challenge, we introduce the \textbf{Template-based Data Generation (TDG)} paradigm. TDG provides a systematic and highly scalable framework for generating a virtually limitless supply of diverse mathematical problems and their corresponding, programmatically verifiable solutions. Our key innovation is the use of a frontier model, GPT-4, to automatically generate \textit{meta-templates}: abstract, parameterized blueprints for problems. This meta-level generation moves beyond simple data augmentation, creating a rich and varied foundation of problem structures that capture a wide spectrum of mathematical concepts and linguistic styles.

Applying this paradigm, we construct and release \textbf{TemplateGSM}, the first installment of our TemplateMath suite. TemplateGSM is a large dataset of over 7 million synthetic grade-school math problems, each accompanied by verified solutions in both code and natural language. The automated generation and rigorous verification pipeline, which leverages programmatic execution, ensures an unprecedented standard of quality and correctness at scale. This resource provides not only clean data for supervised fine-tuning but also a scalable source of feedback for alignment, forming the basis for \textbf{Reinforcement Learning with Verifiable Rewards (RLVR)}. Furthermore, the principles of TDG are generalizable, offering a blueprint for tackling data scarcity in other reasoning-intensive domains such as science, logic, and programming.

Our main contributions are:

\begin{itemize}[leftmargin=*, topsep=0pt, itemsep=0pt] 
\item We introduce TDG, a scalable and principled framework for generating effectively infinite high-quality, domain-specific data. By using a frontier LLM (GPT-4) to generate meta-templates, we establish a new, more powerful approach to data synthesis that systematically overcomes the data scarcity bottleneck.
\item We construct and release \textbf{TemplateGSM}, a dataset of over 7 million math problems with verified solutions. At over 500 times the size of the widely used MATH benchmark, it is the largest dataset of its kind. It provides a critical resource for training and evaluating the next generation of reasoning models.
\item We demonstrate a robust methodology for ensuring correctness at scale. By coupling generation with programmatic execution, our framework provides a binary, verifiable reward signal for every problem. This enables novel training and alignment paradigms such as Reinforcement Learning with Verifiable Rewards (RLVR), moving beyond the limitations of human-annotated preference data.
\item Our work provides a generalizable blueprint for applying TDG to other complex reasoning domains. We believe this methodology will catalyze progress in areas such as scientific problem-solving, legal reasoning, and automated code generation.
\end{itemize}

\section{Template-based Data Generation}

Template-based Data Generation (TDG) is a principled and systematic framework for producing vast quantities of high-quality mathematical problems and their corresponding solutions. To move beyond the limitations of traditional data augmentation, TDG uniquely employs a frontier LLM, GPT-4, to generate \textit{meta-templates}. These are not merely static problem formats but are abstract, parameterized structures that encode deep semantic and logical patterns across diverse problem types and linguistic styles. By instantiating and varying parameters within these sophisticated, GPT-4-generated templates, TDG achieves unprecedented scalability while upholding rigorous quality standards through integrated programmatic verification.

\subsection{Methodology}

The TDG process comprises several key components that collectively yield a high-quality mathematical dataset ready for both supervised training and reinforcement learning.

\subsubsection{Generation of Meta-Templates with LLMs}

We begin by leveraging large language models (LLMs), such as GPT-4, to generate meta-templates \citep{zhang2023meta}. These meta-templates encode the underlying structures of various mathematical problem categories. GPT-4’s advanced language-generation capabilities allow us to produce a wide range of templates, each reflecting different mathematical concepts and problem types. Placeholders for variable components (e.g., names, items, quantities, dates, locations) ensure that the final dataset is both linguistically diverse and contextually rich, contributing to a comprehensive training corpus.

\begin{figure}[ht!]
\centering
\begin{tikzpicture}[
    node distance=0.8cm and 2.0cm,
    llm/.style={rectangle, draw=blue!60, fill=blue!20, thick, minimum width=3cm, minimum height=1cm, align=center},
    process/.style={rectangle, draw=green!60, fill=green!20, thick, minimum width=3cm, minimum height=1cm, align=center},
    data/.style={ellipse, draw=red!60, fill=red!20, thick, minimum width=3cm, minimum height=1cm, align=center},
    decision/.style={diamond, draw=purple!60, fill=purple!20, thick, aspect=2, align=center},
    edge_style/.style={draw=black, thick, ->, >=Latex},
    loop_edge_style/.style={draw=black, thick, ->, >=Latex, dashed},
    label_style/.style={sloped, anchor=south, text=black, font=\small}
]

\node[llm] (gpt4) {LLM Generates\\Meta-Templates};

\node[process, below=of gpt4] (qa_gen) {Q\&A Instantiation};

\node[process, below=of qa_gen] (verification) {Programmatic Verification\\(Code Execution)};

\node[decision, below=of verification] (decision) {Verified?};

\node[data, below=of decision, yshift=-0.5cm] (dataset) {Add to Dataset};

\draw[edge_style] (gpt4) -- (qa_gen);
\draw[edge_style] (qa_gen) -- (verification);
\draw[edge_style] (verification) -- (decision);

\draw[edge_style] (decision) -- node[right]{Yes} (dataset);

\draw[loop_edge_style] (decision.west) to[bend left=45] node[left]{No} (qa_gen.west);

\end{tikzpicture}
\caption{Flowchart illustrating the Template-based Data Generation (TDG) process. An LLM generates meta-templates, which are instantiated into Q\&A pairs. These pairs undergo programmatic verification. The binary outcome (pass/fail) serves as both a quality filter and a verifiable reward signal.}
\label{fig:tdg_flowchart}
\end{figure}

\subsubsection{Generation with Integrated Verification and Verifiable Rewards}

In a single, integrated step, we instantiate specific problems and their corresponding solutions from the GPT-4-produced meta-templates. Each parameter is carefully selected to satisfy problem constraints, ensuring both solvability and validity.

To guarantee correctness and enable advanced training paradigms, we employ a rejection-sampling-based verification process centered on programmatic execution. For each generated problem, its corresponding code-based solution is executed in a sandboxed environment. This verification is objective, scalable, and unambiguous:
\begin{itemize}
    \item A successfully executed code that runs without errors and produces a result is considered verified. The problem-solution pair is accepted.
    \item Any pair that fails, due to runtime errors, incorrect logic, or ill-formed code, is rejected.
\end{itemize}
This binary outcome (success/failure) is crucial. For dataset construction, it acts as a quality filter. For model training, it provides a perfect, high-fidelity reward signal. This mechanism forms the basis for {Reinforcement Learning with Verifiable Rewards (RLVR)}, where a model can be rewarded ($+1$) for generating a verifiably correct solution and penalized ($0$ or $-1$) for an incorrect one, enabling direct optimization of reasoning capabilities without human intervention.

\subsection{Process Flowchart}

An illustrative overview of our TDG method is shown in Figure~\ref{fig:tdg_flowchart}. The flowchart begins with meta-template generation by an LLM (e.g., GPT-4) and concludes with the final dataset. After meta-templates are created, parameters are substituted to instantiate both problems and solutions. The resulting Q\&A pairs then undergo programmatic verification. Any invalid pairs are discarded, and the process iterates until the dataset is populated with validated, high-quality data, with each entry implicitly carrying a verifiable reward.

\subsection{Code Implementation Example}

An illustrative example of our TDG method is presented in Figure~\ref{fig:tdg-example}. The code snippet shows how we generate problems concerning sales over two consecutive months. The meta-template for this scenario, produced by GPT-4, can be varied through parameter substitution to yield many distinct problems. We maintain lists of random terms (e.g., names, items, months, locations) to ensure diversity and realism. This variation prevents repetition and helps models generalize better.

\begin{figure}[ht!]
\centering
\begin{tcolorbox}[width=0.95\textwidth, colback=cyan!2!white, colframe=gray!75!blue]
\small
\begin{lstlisting}[language=Python, basicstyle=\small\ttfamily]
def generate_problem_and_solution_code():
    # Get initial amount and subsequent ratio
    initial_amount, subsequent_ratio = get_params_combination()
    
    # Randomly select terms
    name = random.choice(first_names) + ' ' + random.choice(last_names)
    item = random.choice(items)
    ...
    
    # Construct problem statement
    problem_statement = f"{name} sold {initial_amount} {item} in {month.split(' and ')[0]}, {year} at {place} in {county}. "
    problem_statement += f"In {month.split(' and ')[1]}, they sold {subsequent_ratio*100:.0f}% of the amount sold in the previous month. "
    problem_statement += f"How many {item} did {name} sell in total during {month}?"
    
    # Generate solution code
    sales_var = f"{item.replace(' ', '_')}_sold_in_{month.split(' ')[0]}"
    ratio_var = f"{item.replace(' ', '_')}_ratio"
    total_var = f"total_{item.replace(' ', '_')}"
    solution_code = f"""# Number of {item} sold by {name} in {month.split(' and ')[0]}, {year}
    {sales_var} = {initial_amount}
    # Sales ratio for the next month
    {ratio_var} = {subsequent_ratio}
    # Calculating the amount of {item} sold in {month.split(' and ')[1]}
    subsequent_{sales_var} = {sales_var} * {ratio_var}
    # Calculating the total number of {item} sold during {month}
    {total_var} = {sales_var} + subsequent_{sales_var}
    
    result = {total_var}
    """

    # Execute the solution code for verification
    exec_globals = {}
    exec(solution_code, {}, exec_globals)
    result = round(exec_globals['result'])

\end{lstlisting}
\end{tcolorbox}
\caption{An example of our TDG method. The code demonstrates how variable parameters are used to generate unique problem statements and corresponding solutions based on GPT-4-generated meta-templates, with integrated verification via code execution.}
\label{fig:tdg-example}
\vspace{1ex}
\end{figure}

\subsection{Generated Problem and Solution Example}

To demonstrate TDG’s output, consider this example generated from a GPT-4-created template:

\textbf{Problem:}

\emph{Emily has 15 apples. She buys 3 times more apples and then gives away 5 apples to her friend. How many apples does Emily have now?}

\textbf{Solution:}

Emily starts with 15 apples. She purchases $3$ times more, meaning $15 \times 3 = 45$ apples. This brings her total to $15 + 45 = 60$. After giving away 5 apples, she is left with $60 - 5 = 55$. Thus, Emily has 55 apples remaining.

\textbf{Code-based Solution (The Verifier):}

\begin{lstlisting}[language=Python]
# Initial number of apples Emily has
initial_apples = 15

# Emily buys 3 times more apples
apples_bought = initial_apples * 3

# Total apples after buying more
total_apples = initial_apples + apples_bought

# Emily gives away 5 apples
apples_given_away = 5

# Apples Emily has now
apples_now = total_apples - apples_given_away

result = apples_now  # Emily has 55 apples now
\end{lstlisting}

This Q\&A pair is generated and verified in a single step. The successful execution of the code-based solution confirms its correctness and assigns it a positive verification status.

\subsection{Advantages of TDG}

TDG delivers several notable advantages for generating large-scale mathematical datasets:

\begin{itemize}
\item \textbf{Infinite Scalability}: Through parameter variation in GPT-4-generated templates, TDG can produce effectively infinite data, addressing the extensive training needs of large language models.
\item \textbf{Programmatic Verifiability}: By integrating generation with code execution, every problem-solution pair is confirmed to be correct and consistent. This meticulous, automated verification boosts dataset quality and provides a reliable signal for model training and alignment.
\item \textbf{Rich Diversity}: GPT-4-generated meta-templates encompass a wide range of problem structures and linguistic variants, improving generalization to unseen problem types, and enabling the synthesis of large-scale, high-quality data.
\end{itemize}

\section{TemplateMath Part I: TemplateGSM Dataset}

\subsection{Dataset Construction}

Building upon the powerful TDG paradigm, we have developed \textbf{TemplateGSM}, a foundational dataset comprising over 7 million grade school math problems. As the first installment of the \textbf{TemplateMath} project, TemplateGSM is meticulously designed to foster robust mathematical reasoning. Each problem is paired with both a programmatically verifiable code-based solution and a clear natural language explanation. The problems span a comprehensive range of grade-school-level topics, including arithmetic, fractions, percentages, and basic algebra. Crucially, the generative meta-templates were authored by GPT-4, ensuring an unparalleled diversity of problem structures, narrative contexts, and linguistic expressions. This comprehensive coverage and massive scale make TemplateGSM an essential resource for training and evaluating models across a wide spectrum of problem types and difficulty levels.

\subsection{Dataset Statistics}

The key statistics of the TemplateGSM dataset, summarized in Table~\ref{table:dataset-stats}, underscore its scale and complexity. With 7,473,000 problems generated from 7,473 distinct GPT-4-authored templates, the dataset offers extensive diversity in problem structure and content. To place its scale in perspective, \textbf{TemplateGSM is over 500 times larger than the widely-used MATH benchmark} \citep{hendrycks2021measuring}, providing orders of magnitude more data for training robust models.

\begin{table}[h]
\centering
\caption{Statistics of the TemplateGSM Dataset (o-200k tokenizer)}
\label{table:dataset-stats}
\begin{tabular}{lcc}
\toprule
\textbf{Metric} & \textbf{Value} \\
\midrule
Number of source templates       & 7,473 \\
Total number of problems         & 7,473,000 \\
Problem length range (tokens)    & [18, 636] \\
Code solution length range (tokens)  & [30, 513] \\
Code solution length average (tokens)  & 123.43 $\pm$ 40.82 \\
Natural language solution length range (tokens)  & [1, 1024] \\
Natural language solution length average (tokens)  & 77.87 $\pm$ 33.03 \\
\bottomrule
\end{tabular}
\end{table}

The average lengths for problems and solutions indicate that the dataset provides substantial context and detailed, step-by-step explanations. This richness is highly beneficial for training models not just to find an answer, but to learn the underlying reasoning process. The scale and quality of TemplateGSM also open up new research avenues, such as studying scaling laws for mathematical reasoning and analyzing model failure modes with unprecedented statistical power.

\subsection{Dataset Availability}

\textbf{TemplateGSM} is publicly available and can be accessed at \href{https://huggingface.co/datasets/math-ai/TemplateGSM}{datasets/math-ai/TemplateGSM}. The code used for data generation is also provided at \href{https://github.com/iiis-ai/TemplateMath}{github.com/iiis-ai/TemplateMath}. By sharing both the dataset and the generation code, we enable researchers and practitioners to reproduce our results, extend the dataset, and apply the TDG method to other domains or problem types.

The TemplateGSM dataset is positioned to become a cornerstone resource for pre-training, fine-tuning, and aligning LLMs on mathematical reasoning. By decisively addressing the data scarcity problem, it directly enables the development of more capable and reliable models, a potential already being explored by leading industry labs, as seen with resources like IBM's Granite Language Model~\citep{granite2024granite}. The use of GPT-4 for meta-template generation imbues the dataset with a level of diversity and linguistic naturalness that closely mimics human-authored problems, ensuring that models trained on TemplateGSM can generalize effectively. We anticipate that the TemplateGSM dataset and the broader TemplateMath project will be instrumental in driving fundamental advancements in LLM research focused on reasoning and complex problem-solving.

\section{Related Work}

\paragraph{Mathematical Datasets.} The development of mathematical datasets has been pivotal in advancing LLM research. Early datasets like \textsc{AQUA-RAT} \citep{ling2017program} provided question-answer pairs for arithmetic problems. The \textsc{MATH} dataset \citep{hendrycks2021measuring} established a challenging benchmark with competition-level problems, but its limited size restricts its use for large-scale training. To expand available resources, \citet{paster2023openwebmath} introduced \textsc{OpenWebMath} from web data; however, its utility is often compromised by inconsistent quality and a lack of verified solutions, a gap our verification-centric TDG pipeline explicitly addresses. \textsc{Proof-Pile} \citep{azerbayev2023proofnet} aggregates mathematical texts but lacks structured, executable problem-solution pairs. Similarly, \citet{zhang2025autonomous} uses LLMs as verifiers for data selection, but the resulting dataset remains unstructured. Our work provides a direct solution by generating structured, programmatically verified, and high-quality data at a scale that surpasses these previous efforts.

\paragraph{Training LLMs on Mathematical Tasks.} Base LLMs, despite their broad knowledge \citep{brown2020language, touvron2023llama}, require specialized training for mathematical reasoning. This is typically done via continual pre-training \citep{lewkowycz2022solving, azerbayev2023proofnet} or supervised fine-tuning (SFT) \citep{yu2023metamath, yue2023mammoth, weng2023large}. SFT, while effective, is fundamentally constrained by the availability of high-quality, verified datasets. Our work overcomes this data constraint, providing a massive, reliable corpus needed for effective SFT and for more advanced alignment techniques.

\paragraph{Data Generation and Alignment.} Synthetic data generation has been used to augment training sets \citep{feng2021survey}, with methods like rephrasing \citep{yu2023metamath} or recombination \citep{koncel2015parsing}. However, these methods often lack scalability and rigorous verification. Our TDG approach represents a paradigm shift, offering a systematic engine to generate a virtually infinite number of high-quality problems, each coupled with a verifiable solution. This verifiability connects directly to model alignment techniques like Reinforcement Learning from Human Feedback (RLHF), where models are tuned based on human preferences. TDG enables a more scalable and objective alternative, which we explore in the concept of RLVR.

\section{Conclusion}

We have introduced \textbf{Template-based Data Generation} (\textbf{TDG}), a novel paradigm for generating large-scale, high-quality mathematical datasets. Utilizing TDG, we created \textbf{TemplateGSM}, a dataset of over 7 million grade school math problems, each with a programmatically verifiable solution. The core innovation lies in using GPT-4 to author meta-templates and employing code execution as a scalable, objective verification mechanism. This ensures data quality and provides a reliable foundation for training and evaluation.

By establishing a framework for programmatic verification, we move beyond simple data provision. This work opens the door to advanced training methodologies such as \textbf{Reinforcement Learning with Verifiable Rewards (RLVR)}, offering a more scalable and objective alternative to human-in-the-loop alignment. We are confident that the TDG framework and the TemplateGSM dataset will be instrumental in advancing research in mathematical reasoning. By providing a definitive solution to the data and verification bottleneck, our work lays the groundwork for a new era of large language models capable of deep, reliable, and verifiable reasoning.

\section*{Broader Impact}

The introduction of the TDG paradigm and the release of the TemplateGSM dataset have significant implications for the future of LLMs and open up numerous avenues for future research. We believe our work will catalyze progress in LLM reasoning.

\paragraph{Advancing Mathematical Reasoning via RLVR.} We propose that the most transformative impact of our work lies in enabling Reinforcement Learning with Verifiable Rewards (RLVR). Unlike RLHF, which relies on costly and subjective human feedback, RLVR can leverage the programmatic verifier of TDG as an automated, objective reward function. A language model can be trained to generate a solution (e.g., in code), which is then executed. A successful execution provides a positive reward, while a failure provides a negative reward. This creates a tight, automated feedback loop for improving the model's reasoning abilities at a massive scale. TemplateGSM serves as the perfect environment for this research, and we encourage the community to explore RLVR as a path toward highly reliable mathematical reasoning models.

\paragraph{Generalization to Other Domains.} The TDG framework is not limited to mathematics. Its core principles: LLM-authored meta-templates and programmatic verification, can be readily applied to other domains where correctness can be formally checked. We foresee immediate applications in:
\begin{itemize}
\item  Generating physics, chemistry, or biology problems based on verifiable equations and simulations.
\item Creating datasets of logical puzzles or case-based legal scenarios with verifiable deductive steps.
\item Synthesizing programming challenges where correctness is verified by unit tests.
\end{itemize}

\vspace{5ex}
\bibliographystyle{plainnat}
\bibliography{reference}





\end{document}